\documentclass[10pt,twocolumn,letterpaper]{article}

\usepackage{cvpr}
\usepackage{times}
\usepackage{epsfig}
\usepackage{graphicx}
\usepackage{caption}
\usepackage{amsthm}
\usepackage{amsmath}
\usepackage{amssymb}
\usepackage{bm}
\usepackage{enumitem}
\usepackage{makecell}
\usepackage{wrapfig}
\usepackage{indentfirst}
\usepackage{verbatim}
\usepackage{color}
\usepackage{setspace}

\newcommand{\norm}[1]{\left\lVert#1\right\rVert}
\newcommand{\inp}[2]{\langle#1, #2\rangle}

\renewcommand\thefootnote{}

\usepackage[pagebackref=true,breaklinks=true,letterpaper=true,colorlinks,bookmarks=false]{hyperref}
\hypersetup{linkcolor=[rgb]{0.8551,0.2333,0.2333}}
\hypersetup{citecolor=[rgb]{0.3333,0.2333,0.7551}}

\cvprfinalcopy

\def\cvprPaperID{4133}

\ifcvprfinal\fi
\begin{document}

\title{Decoupled Networks}

\author{\fontsize{10pt}{\baselineskip}\selectfont Weiyang Liu\textsuperscript{1*}, Zhen Liu\textsuperscript{1*}, Zhiding Yu\textsuperscript{2}, Bo Dai\textsuperscript{1}, Rongmei Lin\textsuperscript{3}, Yisen Wang\textsuperscript{1,4}, James M. Rehg\textsuperscript{1}, Le Song\textsuperscript{1,5}\\
\fontsize{11pt}{\baselineskip}\selectfont  \textsuperscript{1}Georgia Institute of Technology\ \ \ \ \textsuperscript{2}NVIDIA\ \ \ \ \textsuperscript{3}Emory University\ \ \ \ \textsuperscript{4}Tsinghua University\ \ \ \ \textsuperscript{5}Ant Financial\\
}

\maketitle
\thispagestyle{empty}

\begin{abstract}
Inner product-based convolution has been a central component of convolutional neural networks (CNNs) and the key to learning visual representations. Inspired by the observation that CNN-learned features are naturally decoupled with the norm of features corresponding to the intra-class variation and the angle corresponding to the semantic difference, we propose a generic decoupled learning framework which models the intra-class variation and semantic difference independently. Specifically, we first reparametrize the inner product to a decoupled form and then generalize it to the decoupled convolution operator which serves as the building block of our decoupled networks. We present several effective instances of the decoupled convolution operator. Each decoupled operator is well motivated and has an intuitive geometric interpretation. Based on these decoupled operators, we further propose to directly learn the operator from data. Extensive experiments show that such decoupled reparameterization renders significant performance gain with easier convergence and stronger robustness.
\end{abstract}

\vspace{-1mm}
\section{Introduction}
Convolutional neural networks have pushed the boundaries on a wide variety of vision tasks, including object recognition~\cite{simonyan2014very,szegedy2015going,he2016deep}, object detection~\cite{girshick2015fast,ren2015faster,redmon2016you}, semantic segmentation~\cite{long2015fully}, etc. A significant portion of recent studies on CNNs focused on increasing network depth and representation ability via improved architectures such as shortcut connections~\cite{he2016deep,huang2017densely} and multi-branch convolution~\cite{szegedy2015going,xie2017aggregated}. Despite these advances, understanding how convolution naturally leads to discriminative representation and good generalization remains an interesting problem.\footnote{\textsuperscript{*}Equal contributions. \ \ \ Email:\{wyliu,liuzhen1994\}@gatech.edu}

\setcounter{footnote}{0}
\renewcommand\thefootnote{\arabic{footnote}}
\begin{figure}[t]
  \centering
  \renewcommand{\captionlabelfont}{\footnotesize}
  \setlength{\abovecaptionskip}{-1.5pt}
  \setlength{\belowcaptionskip}{-14pt}
  \includegraphics[width=2.5in]{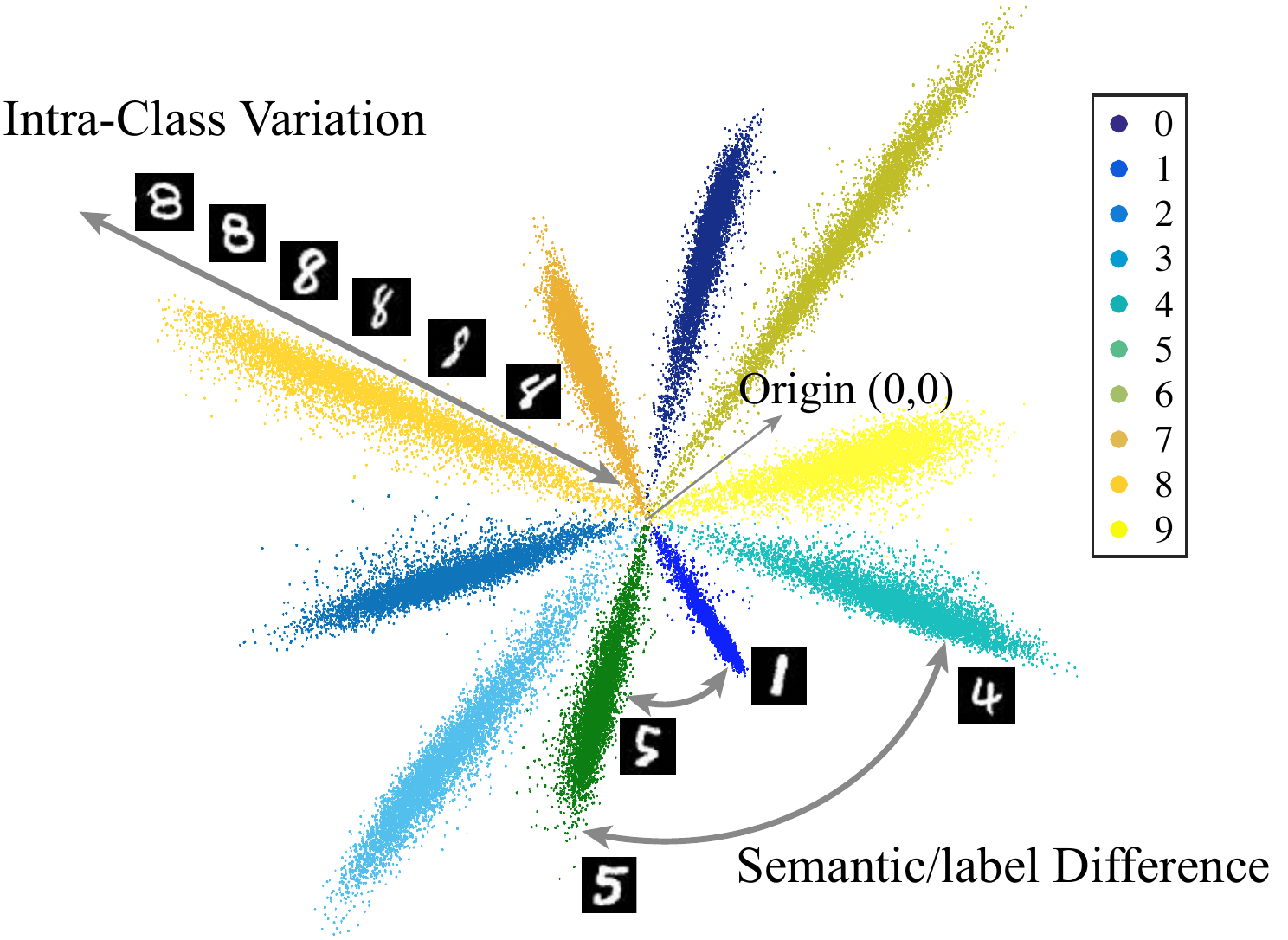}
  \caption{\footnotesize CNN learned features are naturally decoupled. These 2D features are output directly from the CNN by setting the feature dimension as 2.}\label{fig1}
\end{figure}
\par
Current CNNs often encode the similarity between a patch $\bm{x}$ and a kernel $\bm{w}$ via inner product. The formulation of inner product $\thickmuskip=2mu \medmuskip=2mu \langle \bm{w},\bm{x}\rangle=\bm{w}^\top\bm{x}$ couples the semantic difference (\emph{i.e.}, inter-class variation) and the intra-class variation in one unified measure. As a result, when the inner product between two samples is large, one can not tell whether the two samples have large semantic/label difference or have large intra-class variation. In order to better study the properties of CNN representation and further improve existing frameworks, we propose to explicitly decouple semantic difference and intra-class variation\footnote{Although the concepts of semantic difference and intra-class variation often refer to classification, they are extended to convolutions in this paper. Specifically, semantic difference means the pattern similarity between local patch $\bm{x}$ and kernel $\bm{w}$, while intra-class variation refers to the energy of local patch $\bm{x}$ and kernel $\bm{w}$.}. Specifically, we reparametrize the inner product with the norms and the angle, \emph{i.e.}, $\|\bm{w}\|_2\|\bm{x}\|_2\cos(\theta_{(\bm{w},\bm{x})})$. Our direct intuition comes from the the observation in Fig.~\ref{fig1} where angle accounts for semantic/label difference and feature norm accounts for intra-class variation. The larger the feature norm, the more confident the prediction. Such naturally decoupled phenomenon inspires us to propose the decoupled convolution operators. We hope that decoupling norm and angle in inner product can better model the intra-class variation and the semantic difference in deep networks.
\par
On top of the idea to decouple the norm and the angle in an inner product, we propose a novel decoupled network (DCNet) by generalizing traditional inner product-based convolution operators ($\|\bm{w}\|\|\bm{x}\|\cos(\theta_{(\bm{w},\bm{x})})$) to decoupled operators. To this end, we define such operator as multiplication of a function of norms $h(\|\bm{w}\|,\|\bm{x}\|)$ and a function of angle $g(\theta_{(\bm{w},\bm{x})})$. The decoupled operator provides a generic framework to better model the intra-class variation and the semantic difference, and the original CNNs are equivalent to setting $h(\|\bm{w}\|,\|\bm{x}\|)$ as $\|\bm{w}\|\|\bm{x}\|$ and $g(\theta_{(\bm{w},\bm{x})})$ as $\cos(\theta_{(\bm{w},\bm{x})})$. The magnitude function $h(\|\bm{w}\|,\|\bm{x}\|)$ models the intra-class variation while the angular function $g(\theta_{(\bm{w},\bm{x})})$ models the semantic difference.
\par
From the decoupling point of view, the original CNNs make a strong assumption that the intra-class variation can be linearly modeled via the multiplication of norms and the semantic difference is described by the cosine of the angle. However, this modeling approach is not necessarily optimal for all tasks. With the decoupled learning framework, we can either design the decoupled operators based on the task itself or learn them directly from data. The advantages of DCNets are in four aspects. First, DCNets not only allow us to use some alternative functions to better model the intra-class variation and the semantic difference, but they also enable us to directly learn these functions rather than fixing them. Second, with bounded magnitude functions, DCNets can improve the problem conditioning as analyzed in \cite{liu2017deephyper}, and therefore DCNets can converge faster while achieving comparable or even better accuracy than the original CNNs. Third, some instances of DCNets can have stronger robustness against adversarial attacks. We can squeeze the feature space of each class with a bounded $h(\cdot)$, which can bring certain robustness. Last, the decoupled operators are very flexible and architecture-agnostic. They could be easily adapted to any kind of architectures such as VGG~\cite{simonyan2014very}, GoogleNet~\cite{szegedy2015going} and ResNet~\cite{he2016deep}.
\par
Specifically, we propose two different types of decoupled convolution operators: \emph{bounded operators} and \emph{unbounded operators}. We present multiple instances for each type of decoupled operators. Empirically, the bounded operators may yield faster convergence and better robustness against adversarial attacks, and the unbounded operators may have better representational power. These decoupled operators can also be either smooth or non-smooth, which can yield different behaviors. Moreover, we introduce a novel concept - \emph{operator radius} for the decoupled operators. The operator radius describes the critical change of the derivative of the magnitude function $h(\cdot)$ with respect to the input $\|\bm{x}\|$. By jointly learning the operator radius via back-propagation, we further propose \emph{learnable decoupled operators}. Moreover, we show some alternative ways to optimize these operators that improve upon standard back-propagation.
Our contributions can be summarized as:
\par
\begin{itemize}[leftmargin=*,nosep,nolistsep]
    \item Inspired by the observation that CNN-learned features are naturally decoupled, we propose an explicitly decoupled framework to study neural networks.
    \item We show that CNNs make a strong assumption to model the intra-class and inter-class variation, which may not be optimal. By decoupling the inner product, we are able to design more effective magnitude and angular functions rather than the original convolution for different tasks.
    \item In comparison to standard CNNs, DCNets have easier convergence, better accuracy and stronger robustness.
\end{itemize}

\section{Related Works}
\vspace{-0.5mm}
There are an increasing number of works~\cite{wang2018additive,ranjan2017l2,liu2017sphereface,liu2016large,liu2017rethinking,wang2017normface,yuan2017feature,jones2017improving} that focus on improving the classification layer in order to increase the discriminativeness of learned features. \cite{liu2016large} models the angular function for each class differently and defines a more difficult task than classification, improving the network generalization. Built upon \cite{liu2016large}, \cite{liu2017sphereface} further normalizes the weights of the last fully connected layer (\emph{i.e.}, classification layer) and reported improved results on face recognition. \cite{wang2018additive,ranjan2017l2,wang2017normface} normalize the input features before entering the last fully connected layer, achieving promising performance on face recognition. However, these existing works can be viewed as heuristic modifications and are often restricted to the last fully connected layer. In contrast, the decoupled learning provides a more general and systematic way to study the CNNs. In our framework, the previous work can be viewed as proposing a new magnitude function $h(\|\bm{w}\|,\|\bm{x}\|)$ or angular function $g(\theta_{(\bm{w},\bm{x})})$ for the last fully connected layer. For example, normalizing the weights is to let $h(\|\bm{w}\|,\|\bm{x}\|)$ be $\|\bm{x}\|$ and normalizing the input is equivalent to $\thickmuskip=2mu \medmuskip=2mu h(\|\bm{w}\|,\|\bm{x}\|)=\|\bm{w}\|$.
\par
\cite{liu2017deephyper} proposes a deep hyperspherical learning framework which directly makes $\thickmuskip=2mu \medmuskip=2mu h(\|\bm{w}\|,\|\bm{x}\|)$ equal to $1$ such that all the activation outputs only depend on $g(\theta_{(\bm{w},\bm{x})})$. The framework provides faster convergence compared to the original CNNs, but is somehow restricted in the sense that $h(\|\bm{w}\|,\|\bm{x}\|)$ is only allowed to be 1, and therefore can be sub-optimal in some cases. From the decoupling perspective, hyperspherical learning only cares about the semantic difference and aims to compress the intra-class variation to a space that is as small as possible, while the decoupled framework focuses on both. As a non-trivial generalization of \cite{liu2017deephyper}, our decoupled network is a more generic and unified framework to model both intra-class variation and semantic difference, providing the flexibility to design or learn both magnitude function $h(\cdot)$ and angular function $g(\cdot)$.
\vspace{-0.5mm}
\section{Decoupled Networks}
\vspace{-1mm}
\subsection{Reparametrizing Convolution via Decoupling}
\vspace{-0.5mm}
For a conventional convolution operator $f(\cdot,\cdot)$, the output is calculated by the inner product of the input patch $\bm{x}$ and the filter $\bm{w}$ (both $\bm{x}$ and $\bm{w}$ are vectorized into columns):
\begin{equation}
\small
f(\bm{w},\bm{x})=\inp{\bm{w}}{\bm{x}} = \bm{w}^\top\bm{x},
\end{equation}
which can be further formulated as a decoupled form that separates the norm and the angle:
\begin{equation}
\small
f(\bm{w},\bm{x}) =  \norm{\bm{w}}\norm{\bm{x}}cos(\theta_{(\bm{w}, \bm{x})}),
\end{equation}
where $\theta_{(\bm{w}, \bm{x})}$ is the angle between $\bm{x}$ and $\bm{w}$. Our proposed decoupled convolution operator takes the general form of
\begin{equation}
\small
f_d(\bm{w},\bm{x}) = h(\norm{\bm{w}},\norm{\bm{x}})\cdot g(\theta_{(\bm{w}, \bm{x})}),
\end{equation}
which explicitly decouples the norm of $\bm{w},\bm{x}$ and the angle between them. We define $h(\|\bm{w}\|,\|\bm{x}\|)$ as the magnitude function and $g(\theta_{(\bm{w}, \bm{x})})$ as the angular activation function. It is easy to see that the decoupled convolution operator includes the original convolution operator as a special case. As illustrated in Fig.~\ref{fig1}, the semantic difference and intra-class variation are usually decoupled and very suitable for this formulation. Based on the decoupled operator, we propose several alternative ways to model the semantic difference and intra-class variation.

\subsection{Decoupled Convolution Operators}

We discuss how to better model the intra-class variation, and then give a few instances of the decoupled operator.
\vspace{-3mm}
\subsubsection{On Better Modeling of the Intra-class Variation}
\vspace{-1mm}
Hyperspherical learning~\cite{liu2017deephyper} has discussed the modeling of the inter-class variation (\emph{i.e.}, the angular function). The design of angular function $g(\cdot)$ is relatively easy but restricted, because it only takes the angle as input. In contrast, the magnitude function $h(\cdot)$ takes the norm of $\bm{w}$ and the norm of $\bm{x}$ as two inputs, and therefore it is more complicated to design. $\|\bm{w}\|$ is the intrinsic property of a kernel itself, corresponding to the importance of the kernel rather than the intra-class variation of the inputs. Therefore, we tend not to include $\|\bm{w}\|$ into the magnitude function $h(\cdot)$. Moreover, removing $\|\bm{w}\|$ from $h(\cdot)$ indicates that all kernels (or operators) are assigned with equal importance, which encourages the network to make decision based on as many kernels as possible and therefore may make the network generalize better. However, incorporating the kernel importance to the network learning can improve the representational power and may be useful when dealing with a large-scale dataset with numerous categories. By combining $\|\bm{w}\|$ back to $h(\cdot)$, the operators become \emph{weighted decoupled operators}. There are multiple ways of incorporating $\|\bm{w}\|$ back to the magnitude function. We will discuss and empirically evaluate these variants later.
\vspace{-3mm}
\subsubsection{Bounded Decoupled Operators}
\vspace{-1mm}
The output of the bounded operators must be bounded by a finite constant regardless of its input and kernel, namely $\thickmuskip=2mu \medmuskip=2mu |f_d(\bm{w},\bm{x})|\leq c$ where $c$ is a positive constant. For simplicity, we first consider the decoupled operator without the norm of the weights (\emph{i.e.}, $\|\bm{w}\|$ is not included in $h(\cdot)$).
\par
\vspace{0.5mm}
\noindent\textbf{Hyperspherical Convolution.} If we let $\thickmuskip=2mu \medmuskip=2mu h(\norm{\bm{w}},\norm{\bm{x}})=\alpha$, we will have the hyperspherical convolution (SphereConv) with the following decoupled form:
        \begin{equation}
        \small
        f_d(\bm{w},\bm{x}) = \alpha\cdot g(\theta_{(\bm{w},\bm{x})}),
        \end{equation}
where $\thickmuskip=2mu \medmuskip=2mu \alpha>0$ controls the output scale. $g(\theta_{(\bm{w},\bm{x})})$ depends on the geodesic distance on the unit hypersphere and typically outputs value from $-1$ to $1$, so the final output is in $[-\alpha,\alpha]$. Usually, we can use $\thickmuskip=2mu \medmuskip=2mu \alpha=1$, which reduces to SphereConv~\cite{liu2017deephyper} in this case. Geometrically, SphereConv can be viewed as projecting $\bm{w}$ and $\bm{x}$ to a hypersphere and then performing inner product (if $\thickmuskip=2mu \medmuskip=2mu g(\theta)=\cos(\theta)$). Based on \cite{liu2017deephyper}, SphereConv improves the problem conditioning in neural networks, making the network converge better.
\par
\vspace{0.5mm}
\noindent\textbf{Hyperball Convolution.} The hyperball convolution (BallConv) uses $\thickmuskip=2mu \medmuskip=2mu h(\norm{\bm{w}},\norm{\bm{x}})=\alpha\min(\norm{\bm{x}},\rho)/\rho$ as its magnitude function. The specific form of the BallConv is
    \begin{equation}
    \small
        f_d(\bm{w},\bm{x}) = \alpha\cdot\frac{\min(\norm{\bm{x}},\rho)}{\rho}\cdot g(\theta_{(\bm{w},\bm{x})}),
    \end{equation}
where $\rho$ controls the saturation threshold for the input norm $\|\bm{x}\|$ and $\alpha$ scales the output range. When $\|\bm{x}\|$ is larger than $\rho$, then the magnitude function will be saturate and output $\alpha$. When $\|\bm{x}\|$ is smaller than $\rho$, the magnitude function grows linearly with $\|\bm{x}\|$. Geometrically, BallConv can be viewed as projecting $\bm{w}$ to a hypersphere and projecting the input $\bm{x}$ to a hyperball, and then performing the inner product (if $\thickmuskip=2mu \medmuskip=2mu g(\theta)=\cos(\theta)$). Intuitively, BallConv is more robust and flexible than SphereConv in the sense that SphereConv may amplify the $\bm{x}$ with very small $\norm{\bm{x}}$, because $\bm{x}$ with small $\norm{\bm{x}}$ and the same direction as $\bm{w}$ could still produce the maximum output. It makes SphereConv  sensitive to perturbations to $\bm{x}$ with small norm. In contrast, BallConv will not have such a problem, because the multiplicative factor $\norm{\bm{x}}$ can help to decrease the output if $\norm{\bm{x}}$ is small. Moreover, small $\norm{\bm{x}}$ indicates that the local patch is not informative and should not be emphasized. In this sense, BallConv is better than SphereConv. In terms of convergence, the BallConv can still help the network convergence because its output is bounded with the same range as SphereConv.

\vspace{0.5mm}
\noindent\textbf{Hyperbolic Tangent Convolution.} We present a smooth decoupled operator with bounded output called hyperbolic tangent convolution (TanhConv). The TanhConv uses a hyperbolic tangent function to replace the step function in the BallConv and can be formulated as
    \begin{equation}
    \small
        f_d(\bm{w},\bm{x}) = \alpha\tanh\big(\frac{\norm{\bm{x}}}{\rho}\big)\cdot g(\theta_{(\bm{w},\bm{x})}),
    \end{equation}
where $\tanh(\cdot)$ denotes the hyperbolic tangent function and $\rho$ is parameter controlling the decay curve. The TanhConv can be viewed as a smooth version of BallConv, which not only shares the same advantages as BallConv but also has more convergence gain due to its smoothness~\cite{clevert2015fast}.
\begin{figure}[t]
  \centering
  \renewcommand{\captionlabelfont}{\footnotesize}
  \setlength{\abovecaptionskip}{3pt}
  \setlength{\belowcaptionskip}{-10pt}
  \includegraphics[width=3.3in]{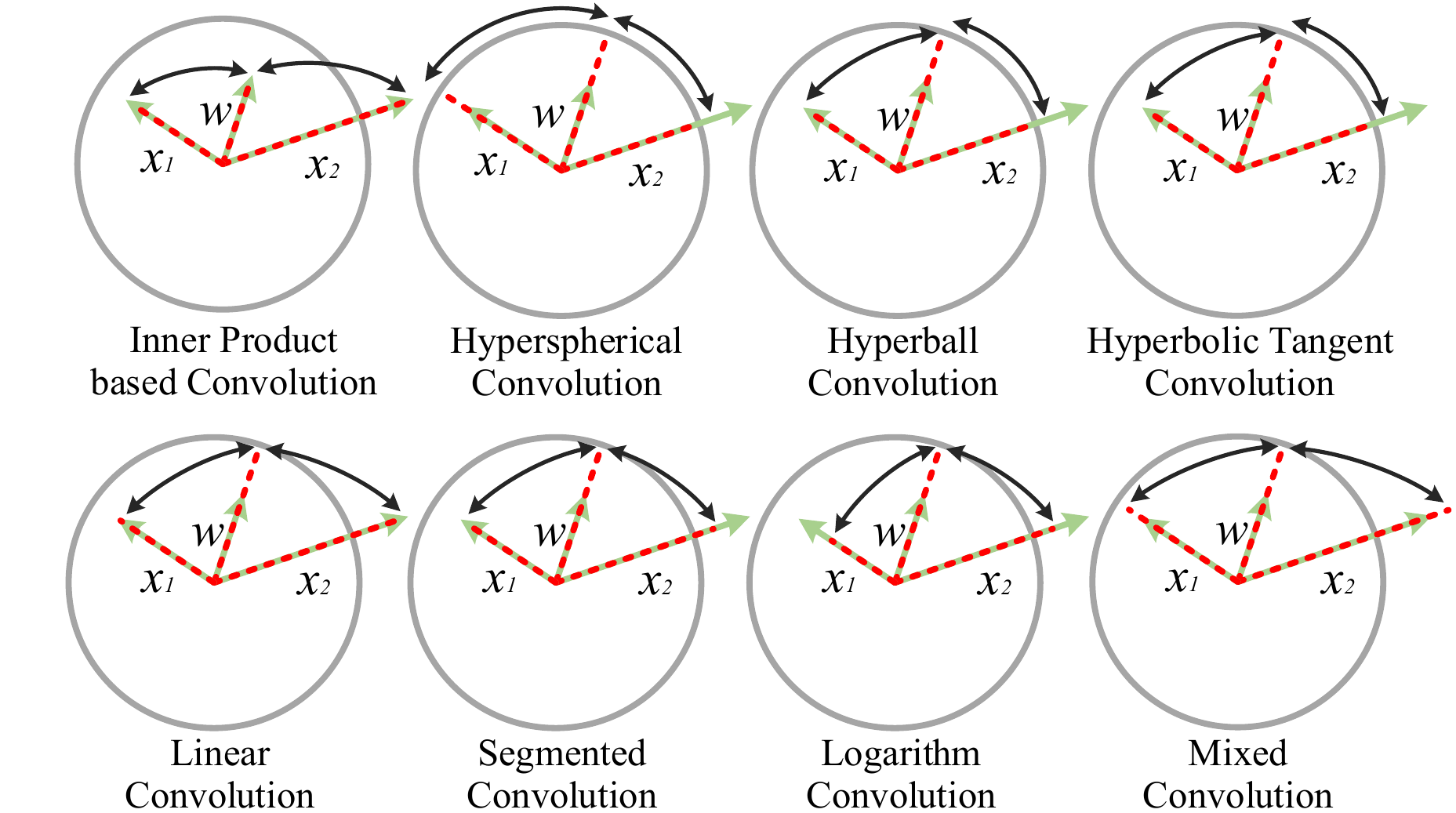}
  \caption{\footnotesize Geometric interpretations for decoupled convolution operators. Green denotes the original vectors, and red denotes the projected vectors. }\label{geo}
\end{figure}
\vspace{-3mm}
\subsubsection{Unbounded Decoupled Operators}
\vspace{-1mm}
\noindent\textbf{Linear Convolution.} One of the simplest unbounded decoupled operators is the linear convolution (LinearConv):
    \begin{equation}
    \small
        f_d(\bm{w},\bm{x}) = \alpha\norm{\bm{x}}\cdot g(\theta_{(\bm{w},\bm{x})}),
    \end{equation}
where $\alpha$ controls the output scale. LinearConv differs the original convolution in the sense that it projects the weights to a hypersphere and has a parameter to control the slope.
\par
\vspace{0.5mm}
\noindent\textbf{Segmented Convolution.} We propose a segmented convolution (SegConv) which takes the following form:
\begin{equation}
\footnotesize
f_d(\bm{w},\bm{x})=
\left\{
{\begin{array}{*{20}{l}}
{\alpha\norm{\bm{x}}\cdot g(\theta_{(\bm{w},\bm{x})}),\ \ \ \ 0\leq\norm{\bm{x}}\leq\rho}\\
{(\beta\norm{\bm{x}}+\alpha\rho-\beta\rho)\cdot g(\theta_{(\bm{w},\bm{x})}),\ \ \ \ \rho<\norm{\bm{x}}}
\end{array}} \right.,
\end{equation}
where $\alpha$ controls the slope when $\thickmuskip=2mu \medmuskip=2mu \|\bm{x}\|\leq \rho$ and $\beta$ controls the slope when $\thickmuskip=2mu \medmuskip=2mu \|\bm{x}\|>\rho$. $\rho$ is the change point of the gradient of the magnitude function w.r.t. $\|\bm{x}\|$. SegConv is a flexible multi-range linear function corresponding to $\|\bm{x}\|$. Both LinearConv and BallConv are special cases of SegConv.
\par
\vspace{0.5mm}
\noindent\textbf{Logarithm Convolution.} We present another smooth decoupled operator with unbounded output, logarithm convolution (LogConv). LogConv uses a Logarithm function for the norm of the input $\|\bm{x}\|$ and can be formulated as
    \begin{equation}
    \small
        f_d(\bm{w},\bm{x}) = \alpha\log(1+\norm{\bm{x}})\cdot g(\theta_{(\bm{w},\bm{x})}),
    \end{equation}
where $\alpha$ controls the base of logarithm and is used to adjust the curvature of the logarithm function.
\par
\vspace{0.5mm}
\noindent\textbf{Mixed Convolution.} Mixed convolution (MixConv) combines multiple decoupled convolution operators and enjoys better flexibility. Because the mixed convolution has many possible combinations, we only consider the additive combination of LinearConv and LogConv as an example:
    \begin{equation}
    \small
        f_d(\bm{w},\bm{x}) = \big(\alpha\norm{\bm{x}}+\beta\log(1+\norm{\bm{x}})\big)\cdot g(\theta_{(\bm{w},\bm{x})}),
    \end{equation}
which combines LogConv and LinearConv, becoming more flexible than both original operators.
\vspace{-3mm}
\subsubsection{Properties of Decoupled Operators}
\vspace{-1mm}
\noindent\textbf{Operator Radius.} Operator radius is defined to describe the gradient change point of the magnitude function. Operator radius differentiates two stages of the magnitude function. The two stages usually have different gradient ranges and therefore behave differently during optimization. We let $\rho$ denote the operator radius in each decoupled operator. For BallConv, when $\|\bm{x}\|$ is smaller than $\rho$, the magnitude function will be activated and it will grow with $\|\bm{x}\|$ linearly. When $\|\bm{x}\|$ is larger than $\rho$, then the magnitude function will be deactivated and output a constant. For SegConv, $\thickmuskip=2mu \medmuskip=2mu \|\bm{x}\|=\rho$ is the change point of the magnitude function's slope. The operator radius of some decoupled operators (SphereConv, LinearConv, LogConv) is defined to be zero, indicating that they have no operator radius. The decoupled operator with non-zero operator radius is similar to a gated operator where $\thickmuskip=2mu \medmuskip=2mu \|\bm{x}\|=\rho$ serves as the switch.

\vspace{0.5mm}
\noindent\textbf{Boundedness.} The Boundedness of a decoupled operator may affect its convergence speed and robustness. \cite{liu2017deephyper} shows that using a bounded operator can improve the convergence due to two reasons. First, bounded operators lead to better problem conditioning in training a deep network via stochastic gradient descent. Second, bounded operators make the variance of the output small and partially address the internal covariate shift problem. The bounded operators can also constrain the Lipschitz constant of a neural network, making the entire network more smooth. The Lipschitz constant of a neural network is shown to be closely related to its robustness against adversarial perturbation~\cite{hein2017formal}. In contrast, the unbounded operators may have stronger approximation power and flexibility than the bounded ones.

\vspace{0.5mm}
\noindent\textbf{Smoothness.} The smoothness of the magnitude function is closely related to the approximation ability and the convergence behavior. In general, using a smooth magnitude function could have better approximation rate~\cite{mhaskar1994choose} and may also lead to more stable and faster convergence~\cite{clevert2015fast}. However, a smooth magnitude function may also be more computationally expensive, since it could be more difficult to approximate a smooth function with polynomials.

\subsection{Geometric Interpretations}\label{geo_sec}
All the decoupled convolution operators have very clear geometric interpretations, as illustrated in Fig.~\ref{geo}. Because all decoupled operators normalize the kernel weights, all the weights are already on the unit hypersphere. SphereConv also projects the input vector $\bm{x}$ on the unit hypersphere and then computes the similarity between $\bm{w}$ and $\bm{x}$ based on the geodesic distance on the hypersphere (multiplied by a scaling factor $\alpha$). Therefore, its output is bounded from $-\alpha$ to $\alpha$ and only depends on the directions of $\bm{w}$ and $\bm{x}$ (suppose $g(\theta_{(\bm{w}, \bm{x})})$ is in the range of $[-1,1]$).
\par
BallConv first projects the input vector $\bm{x}$ to a hyperball and then computes the similarity based on the projected $\bm{x}$ inside the hyperball and the normalized $\bm{w}$ on surface of the hyperball. Specifically, BallConv projects $\bm{x}$ to the hypersphere if $\thickmuskip=2mu \medmuskip=2mu \norm{\bm{x}}>\rho$. TanhConv is a smoothed BallConv and has similar geometric interpretation, but TanhConv is differentible everywhere and has soft boundary around the operator radius $\thickmuskip=2mu \medmuskip=2mu \|\bm{x}\|=\rho$. TanhConv can be viewed as performing projection to a soft hyperball.
\par
SegConv is more flexible than both SphereConv and BallConv. By using certain parameters, SegConv can reduce to either SphereConv or BallConv. SegConv essentially adjusts the norm of the input $\bm{x}$ with a multi-range linear multiplicative factor. Geometrically, such a factor will either push the vector close to the hypersphere or away from the hypersphere depending on the selection of $\alpha$ and $\beta$. For example, we consider the case where $\thickmuskip=2mu \medmuskip=2mu \alpha=1$ and $\thickmuskip=2mu \medmuskip=2mu 0<\beta<1$. When $\thickmuskip=2mu \medmuskip=2mu \|\bm{x}\|\leq \rho$, the magnitude function $h(\cdot)$ in SegConv will directly output $\|\bm{x}\|$. When $\thickmuskip=2mu \medmuskip=2mu \|\bm{x}\|> \rho$, $h(\cdot)$ in SegConv will output a value smaller than $\|\bm{x}\|$, as shown in Fig.~\ref{geo}.
\par
LinearConv is the simplest unbounded operator and its magnitude function grows linearly with $\|\bm{x}\|$. When $\thickmuskip=2mu \medmuskip=2mu \alpha=1$, the magnitude function $h(\cdot)$ in LinearConv simply outputs $\|\bm{x}\|$, which does not perform any projection.
\par
LogConv use a logarithm function to transform the norm of the input $\bm{x}$. After such nonlinear transformation on $\bm{x}$, LogConv computes similarity based on the transformed input $\bm{x}$ and the normalized weights on a hypersphere.

\begin{figure}[t]
  \centering
  \renewcommand{\captionlabelfont}{\footnotesize}
  \setlength{\abovecaptionskip}{3pt}
  \setlength{\belowcaptionskip}{-10pt}
  \includegraphics[width=3.4in]{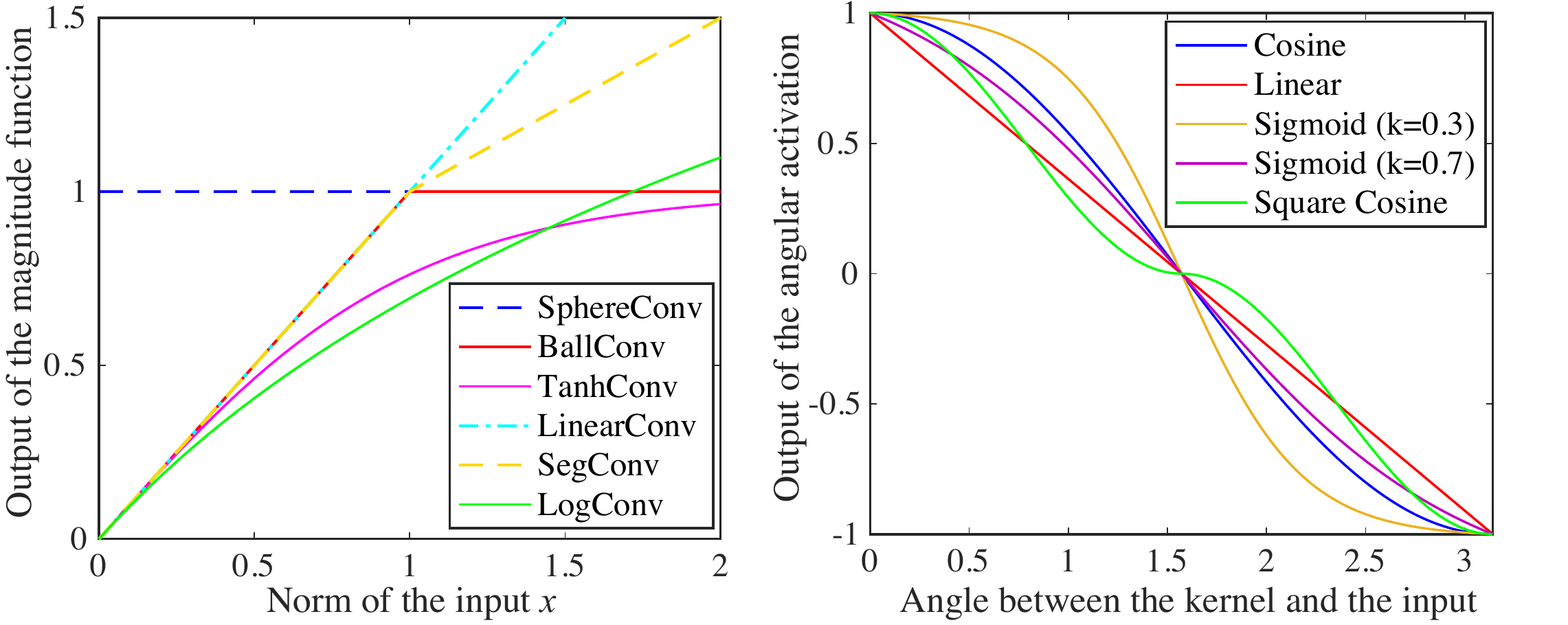}
  \caption{\footnotesize Magnitude function ($\thickmuskip=2mu \medmuskip=2mu \rho=1$) and angular activation function. }\label{angle}
\end{figure}

\subsection{Design of the Angular Activation Function}
The design of the angular function $g(\theta_{(\bm{w},\bm{x})})$ mostly follows the deep hyperspherical learning~\cite{liu2017deephyper}. We use four different types of $g(\theta_{(\bm{w},\bm{x})})$ in this paper. The linear angular activation is defined as
\begin{equation}
    \small
    g(\theta_{(\bm{w},\bm{x})})=-\frac{2}{\pi}\theta_{(\bm{w}, \bm{x})}+1,
\end{equation}
whose output grows linearly with the angle $\theta_{(\bm{w}, \bm{x})}$. The cosine angular activation is defined as
\begin{equation}
    \small
\thickmuskip=2mu \medmuskip=2mu g(\theta_{(\bm{w},\bm{x})})=\cos(\theta_{(\bm{w}, \bm{x})}),
\end{equation}
which is also used by the original convolution operator. Moreover, the sigmoid angular activation is defined as
\begin{equation}
    \small
    g(\theta_{(\bm{w},\bm{x})}) =\frac{1+\exp(-\frac{\pi}{2k})}{1-\exp(-\frac{\pi}{2k})} \cdot \frac{1-\exp({\frac{\theta_{(\bm{w},\bm{x})}}{k}-\frac{\pi}{2k}})}{1+\exp({\frac{\theta_{(\bm{w},\bm{x})}}{k}-\frac{\pi}{2k}})},
\end{equation}
where $k$ controls the curvature. Additionally, we also propose a square cosine angular activation function:
\begin{equation}
    \small
    g(\theta_{(\bm{w},\bm{x})})=\textnormal{sign}(\cos(\theta))\cdot\cos^2(\theta),
\end{equation}
which can encourage a degree of angular margin near the decision boundary and may improve network generalization. In addition to fixing these angular activations prior to training, we can also jointly learn the parameter $k$ in the sigmoid activation using back-propagation, which is a learnable angular activation~\cite{liu2017deephyper}. Fig.~\ref{angle} shows the curves of these angular activation functions.

\subsection{Weighted Decoupled Operators}
All the decoupled operators we have discussed normalize the kernel weights $\bm{w}$ and the magnitude functions do not take the weights into consideration. Although empirically we find that the standard decoupled operators work better than the weighted ones in most cases, we still consider weighted decoupled operators, which incorporate $\|\bm{w}\|$ into the magnitude function, in order to improve the operator's flexibility. We propose two straightforward ways to combine $\|\bm{w}\|$: linear and nonlinear.
\par
\vspace{0.5mm}
\noindent\textbf{Linearly Weighted Decoupled Operator.} Similar to the original inner produce-based convolution, we can directly multiply the norm of weights into the magnitude function, which makes the decoupled operators linearly weighted. For example, SphereConv will become $f_d(\bm{w},\bm{x}) = \alpha\|\bm{w}\|\cdot g(\theta_{(\bm{w},\bm{x})})$. Notably, linearly weighted LinearConv will become the original inner produce-based convolution.
\par
\vspace{0.5mm}
\noindent\textbf{Nonlinearly Weighted Decoupled Operator.} Compared to linearly weighted decoupled operators, the norm of the weights are incorporated into the magnitude function in a nonlinear way. Taking TanhConv as an example, we could formulate the nonlinearly weighted TanhConv as
\begin{equation}\label{eq:nwdov1}
\small
    f_d(\bm{w},\bm{x}) = \alpha\tanh(\frac{1}{\rho}\norm{\bm{x}}\cdot\norm{\bm{w}})\cdot g(\theta_{(\bm{w},\bm{x})}).
\end{equation}
We can also formulate the nonlinearly weighted TanhConv in an alternative way:
\begin{equation}\label{eq:nwdov2}
\small
f_d(\bm{w},\bm{x}) = \alpha\tanh(\frac{1}{\rho}\norm{\bm{w}})\cdot\tanh(\frac{1}{\rho}\norm{\bm{x}})\cdot g(\theta_{(\bm{w},\bm{x})}).
\end{equation}
The first nonlinearly weighted formulation couples $\|\bm{x}\|$ and $\|\bm{w}\|$ by multiplication and then perform a nonlinear transformation, while the second one performs nonlinear transformations separately for $\|\bm{x}\|$ and $\|\bm{w}\|$, and then multiplies them. In practice, the linearly weighted operators are favored over nonlinearly weighted ones due to the simplicity.
\subsection{Learnable Decoupled Operators}
Because our decoupled operators usually have hyperparameters, we usually need to do cross-validation in order to choose suitable parameters, which is time-consuming and sub-optimal. To address this, we can learn these parameters jointly with network weight training via back-propagation.  We propose learnable decoupled operators which perform hyperparameter learning with $h(\cdot)$ and $g(\theta_{(\bm{w},\bm{x})})$. For example, \cite{liu2017deephyper} proposed to learn the hyperparameters of sigmoid angular function. By making both $h(\|\bm{w}\|,\|\bm{x}\|)$ and $g(\theta_{(\bm{w},\bm{x})})$ learnable, we can greatly enhance the representational power and flexibility.
\par
However, making the decoupled operators too flexible (\emph{i.e.}, too many learnable parameters) may require a prohibitive amount of training data to achieve good generalization. In order to achieve an effective trade-off, we only investigate learning the operator radius $\rho$ via back-propagation during the network training.

\section{Improving the Optimization for DCNets}
\vspace{-1mm}
We propose several tricks to improve the optimization of DCNets and enable DCNets to converge to a better local minima. More analysis and discussion of weight projection and weight gradients are provided in Appendix~\textcolor[rgb]{0.7551,0.2333,0.3333}{G}.

\subsection{Weight Projection}
\vspace{-1mm}
The forward pass of DCNets is not dependent on the norm of the weights $\|\bm{w}\|$, because the decoupled operators take the normalized weights as input. However, $\|\bm{w}\|$ will significantly affect the backward pass. Taking SphereConv as an example, we compute the gradient w.r.t. $\bm{w}$:
\vspace{-1mm}
\begin{equation}\label{eq:gradw}
    \small
    \frac{\partial}{\partial \bm{w}}\big( \hat{\bm{w}}^\top\hat{\bm{x}} \big) =
    \frac{\hat{\bm{x}} - \hat{\bm{w}}^\top\hat{\bm{x}}\cdot\hat{\bm{w}}}{\norm{\bm{w}}},
\end{equation}
\par
\noindent where $\thickmuskip=2mu \medmuskip=2mu \hat{\bm{w}} = \bm{w}/\norm{\bm{w}}$ and $\thickmuskip=2mu \medmuskip=2mu \hat{\bm{x}} = \bm{x}/\norm{\bm{x}}$.
In comparison, $\|\bm{w}\|$ will not affect the gradient w.r.t $\bm{w}$ in inner product. From Eq.~\eqref{eq:gradw}, large $\|\bm{w}\|$ can make the gradients very small so that the backward pass is not able to update the weights effectively. To address this issue, we propose \emph{weight projection} to control the norm of the weights. Weight projection performs $\thickmuskip=2mu \medmuskip=2mu \bm{w}\leftarrow s\cdot \hat{\bm{w}}$ every certain number of iterations where $\leftarrow$ denotes the replacement operation. $s$ is a positive constant which controls the norm of the gradient (we use $\thickmuskip=2mu \medmuskip=2mu s=1$ in our experiments). In general, larger $s$ leads to smaller gradients. Note that, weight projection cannot be used in the weighted decoupled operators, because $\|\bm{w}\|$ will affect the forward pass. We can only apply weight projection to our standard decoupled operators.
\subsection{Weighted Gradients}
\vspace{-0.7mm}
From Eq.~\eqref{eq:gradw}, we observe that we could simply multiply $\|\bm{w}\|$ to Eq.~\eqref{eq:gradw} to eliminate the effect of $\|\bm{w}\|$ on the backward pass. We update the weights with the following:
\vspace{-1mm}
\begin{equation}\label{eq:gradwnew}
    \small
    \Delta \bm{w}=\norm{\bm{w}}\cdot\frac{\partial}{\partial \bm{w}}\big( \hat{\bm{w}}^\top\hat{\bm{x}} \big) =
    \hat{\bm{x}} - \hat{\bm{w}}^\top\hat{\bm{x}}\cdot\hat{\bm{w}},
\end{equation}
which does not depend on $\|\bm{w}\|$ and is called \emph{weighted gradients}. Using the proposed weighted gradients for back-propagation, we can also prevent the gradients from being affected by the norm of the weights.
\vspace{-0.3mm}
\subsection{Pretraining as a Better Initialization}\label{finetune}
\vspace{-0.7mm}
We find that DCNets may sometimes be trapped into a bad local minima and yield a less competitive accuracy while trained on  large-scale datasets (\emph{e.g.}, ImageNet). Because the decoupled operators have stronger nonlinearity, its loss landscape may be more complex than the original convolution. The most straightforward way to improve the optimization is to use a better initialization. To this end, we use a CNN model that has the same structure and is pretrained on the same training set to initialize the DCNet.

\section{Discussions}
\vspace{-1.5mm}
\noindent\textbf{Why Decoupling?} Decoupling the intra-class and inter-class variation gives us the flexibility to design better models that are more suitable for a given task. Inner product-based convolution is computationally attractive but not necessarily optimal. The original convolution makes an assumption that the intra-class and inter-class variations are modeled by $\thickmuskip=2mu \medmuskip=2mu h(\|\bm{w}\|,\|\bm{x}\|)=\|\bm{w}\|\|\bm{x}\|$ and $\thickmuskip=2mu \medmuskip=2mu g(\theta)=\cos(\theta)$, respectively. Such assumptions may not be optimal. $h(\cdot)$ and $g(\cdot)$ can be task-driven in our novel decoupled framework.
\par
\vspace{0.5mm}
\noindent\textbf{Flexibility of Decoupled Operators.} There are numerous design options for the magnitude and angular function. The original convolution can be viewed as a special decoupled operator. Moreover, we can parametrize a decoupled operator with a few learnable parameters and learn them via back-propagation. However, there is a delicate tradeoff between the size of the training data, the generalization of the network and the flexibility of the decoupled operator. Generally, given a large enough dataset, the network generalization improves with more learnable parameters.
\par
\vspace{0.5mm}
\noindent\textbf{A Unified Learning Framework for CNNs.} The decoupled formulation provides a unified learning framework for CNNs. Consider a standard CNN with ReLU, we write the convolution and ReLU as $\thickmuskip=2mu \medmuskip=2mu \max(0,\|\bm{w}\|\|\bm{x}\|\cos(\theta))$ which can be written as $\thickmuskip=2mu \medmuskip=2mu \|\bm{w}\|\|\bm{x}\|\cdot\max(0,\cos(\theta))$. Such formulation can be viewed as a decoupled operator where $\thickmuskip=2mu \medmuskip=2mu h(\|\bm{w}\|,\|\bm{x}\|)=\|\bm{w}\|\|\bm{x}\|$ and $\thickmuskip=2mu \medmuskip=2mu g(\theta)=\max(0,\cos(\theta))$. We can jointly consider the convolution operator and nonlinear activation in the decoupled framework. It is possible to learn one single function $g(\cdot)$ that represents both angular activation and the nonlinearity, which is why the square cosine angular activation works well without ReLU.
\par
\vspace{0.5mm}
\noindent\textbf{Network Regularization.} In most instances of  DCNets, the $\ell_2$ weight decay is no longer suitable. \cite{liu2017deephyper} uses an orthonormal constraint $\thickmuskip=2mu \medmuskip=2mu \|\bm{W}^\top\bm{W}-\bm{I}\|_F^2$ to regularize the network, where $\bm{W}$ is the weight matrix whose columns are the kernel weights and $\bm{I}$ is identity matrix. We also
propose an orthogonal constraint $\thickmuskip=2mu \medmuskip=2mu \|\bm{W}^\top\bm{W}-\mathop{\mathrm{diag}}(\bm{W}^\top\bm{W})\|_F^2$.
\par
\vspace{0.5mm}
\noindent\textbf{Network Architecture.}
Due to the non-linear nature of DCNet, the performance of specific $h(\cdot)$ and $g(\cdot)$ is dependent on the choice of architecture. An interesting challenge for future work is to inverstigate the link between the architecture and the choise of $h(\cdot)$ and $g(\cdot)$.

\vspace{-2.5mm}
\section{Experiments and Results}
\vspace{-1.5mm}
\noindent\textbf{General.} We evaluate both accuracy and robustness of DCNets on objection recognition.
For all decoupled operators, we use the standard softmax loss if not otherwise specified.
\par
\noindent\textbf{Training.} The architecture for each task and the training details are given in Appendix~\textcolor[rgb]{0.7551,0.2333,0.3333}{A}. For CIFAR, the network is trained by ADAM with 128 batch size. The learning rate starts from $0.001$. For ImageNet, we use SGD with momentum $0.9$ and batch size $40$. The learning rate starts from $0.1$. For adversarial attacks, the networks are trained by ADAM. All learning rates are divided by $10$ when the error plateaus.
\par
\noindent\textbf{Implementation Details.} For all decoupled operators that have non-zero operator radius (\emph{i.e.}, $\thickmuskip=2mu \medmuskip=2mu \rho\neq 0$), we will learn the operator radius from the training data via back-propagation. More details are provided in Appendix~\textcolor[rgb]{0.7551,0.2333,0.3333}{B}.

\vspace{-0.5mm}
\subsection{Object Recognition}
\vspace{-0.5mm}
\subsubsection{CIFAR-10 and CIFAR-100}
\vspace{-1.5mm}
\noindent\textbf{Weighted Decoupled Operators.}
We first compare the weighted decoupled operators and the standard ones. Because the weights are incorporated into the forward pass in the weighted decoupled operators, the optimization tricks like weight projection and weighted gradients are not applicable. Therefore, the weighted operators simply use the conventional gradients to perform back-propagation. For standard decoupled operators, we show the results using standard optimization, weight projection and weight gradients. From the results of TanhConv in Table~\ref{weighted_op}, weighted decoupled operators do not show obvious advantages.
\begin{table}[h]
\vspace{-1.5mm}
\footnotesize
\renewcommand{\captionlabelfont}{\footnotesize}
  \setlength{\abovecaptionskip}{2pt}
  \setlength{\belowcaptionskip}{-8pt}
\centering
\begin{tabular}{c || c }
  \hline
 Method & Error \\
 \hline\hline
 Linearly Weighted Decoupled Operator                               & 22.95 \\
 Nonlinearly Weighted Decoupled Operator (Eq.~\eqref{eq:nwdov1})    & 23.03 \\
 Nonlinearly Weighted Decoupled Operator (Eq.~\eqref{eq:nwdov2})    & 23.38 \\
 Decoupled Operator (Standard Gradients) & 23.09 \\
 Decoupled Operator (Weight Projection)  & \textbf{21.17} \\
 Decoupled Operator (Weighted Gradients)  & 21.45  \\
 \hline
\end{tabular}
\caption{\footnotesize Evaluation of weighted operators (TanhConv) on CIFAR-100.}
\label{weighted_op}
\end{table}
\par
\vspace{0.5mm}
\noindent\textbf{Optimization Tricks.} We propose weight projection and weighted  gradients to facilitate the optimization of DCNets. These two tricks essentially amplify the original gradient and make the backward update more effective. From Table~\ref{weighted_op}, we observe that both weight projection and weighted gradients work much better than the competing methods.

\begin{table}[h]
\vspace{-1.5mm}
\footnotesize
  \setlength{\abovecaptionskip}{2pt}
  \setlength{\belowcaptionskip}{-8pt}
\centering
\renewcommand{\captionlabelfont}{\footnotesize}
\begin{tabular}{c || c c c}
  \hline
  Method & Linear & Cosine & Sq. Cosine \\
 \hline\hline
 CNN Baseline      &  -          & 35.30             &  -    \\
 LinearConv        & 33.39       & 31.76             & N/C \\
 TanhConv          & \textbf{32.88}       & 31.88                & \textbf{34.26} \\
 SegConv           & 34.69       & \textbf{30.34}    & N/C \\
 \hline
\end{tabular}
\caption{\footnotesize Testing error (\%) of plain CNN-9 without BN on CIFAR-100. ``N/C'' indicates that the model can not converge. ``-'' denotes no result. The results of different columns belong to different angular activation.}
\label{table:bn}
\end{table}

\par
\vspace{0.5mm}
\noindent\textbf{Learning without Batch Normalization.} Batch Normalization (BN)~\cite{ioffe2015batch} is usually crucial for training a well-performing CNN, but the results in Table~\ref{table:bn} show that our decoupled operators can perform much better than the original convolution even without BN.
\par
\vspace{0.5mm}
\noindent\textbf{Learning without ReLU.} Our decoupled operators naturally have strong nonlinearity, because our decoupled convolution is no longer a linear matrix multiplication. In Table~\ref{table:cifar100-vgg}, square cosine angular activation works extremely well in plain CNN-9, even better than the networks with ReLU. The results show that using suitable $h(\cdot)$ and $g(\cdot)$ can lead to significantly better accuracy than the baseline CNN with ReLU, even if our DCNet does not use ReLU at all.

\begin{table}[h]
\vspace{-1mm}
\footnotesize
  \setlength{\abovecaptionskip}{2pt}
  \setlength{\belowcaptionskip}{-11pt}
\centering
\renewcommand{\captionlabelfont}{\footnotesize}
\begin{tabular}{c || c c c c c}
  \hline
  Method  & \makecell{Cosine \\ w/o ReLU} & \makecell{Sq. Cosine \\ w/o ReLU} & \makecell{Cosine \\ w/ ReLU} & \makecell{Sq. Cosine \\ w/ ReLU}  \\
 \hline\hline
 Baseline  & 58.24          &  -             & 26.01 & -  \\
 SphereConv    & 33.31          & 25.90          & 26.00 & 26.97 \\
 BallConv      & \textbf{31.81} & 25.43          & 25.18 & 26.48 \\
 TanhConv      & 32.27          & 25.27          & 25.15 & 26.94\\
 LinearConv    & 36.49          & 24.36          & \textbf{24.81} & 25.14 \\
 SegConv       & 33.57          & \textbf{24.29} & 24.96 & \textbf{25.04} \\
 LogConv       & 33.62          & 24.91          & 25.17 & 25.85 \\
 MixConv       & 33.46          & 24.93          & 25.27 & 25.77\\
 \hline
\end{tabular}
\caption{\footnotesize Testing error rate (\%) of plain CNN-9 on CIFAR-100. Note that, BN is used in all compared models. Baseline is the original plain CNN-9.}
\label{table:cifar100-vgg}
\end{table}

\vspace{0.5mm}
\noindent\textbf{Comparison among Different Decoupled Operators.} We compare different decoupled operators on both plain CNN-9 and ResNet-32. All the compared decoupled operators are unweighted and use weight projection during optimization. The standard softmax loss and BN are used in all networks. For plain CNN-9, we compare the case with and without ReLU. The results in Table~\ref{table:cifar100-vgg} show that DCNets significantly outperform the baseline. In particular, our DCNet with SegConv and square cosine can achieve 24.29\% even without ReLU, which is even better than the networks with ReLU. For ResNet-32, our DCNets also consistently outperform the baseline with a considerable margin. The results further verify that the intra-class and inter-class variation assumptions of the original CNN are not optimal.

\begin{table}[h]
\vspace{-1.2mm}
\footnotesize
  \setlength{\abovecaptionskip}{2pt}
  \setlength{\belowcaptionskip}{-8pt}
\centering
\renewcommand{\captionlabelfont}{\footnotesize}
\begin{tabular}{c || c c c}
  \hline
  Method & Linear & Cosine & Sq. Cosine \\
 \hline\hline
 ResNet Baseline      &  -   & 26.69  &  -    \\
 SphereConv   & 21.79 & 21.44 & 24.40 \\
 BallConv     & 21.44 & 21.12 & 24.31 \\
 TanhConv     & 21.6 & 21.17 & 24.77 \\
 LinearConv    & 21.09 & 22.17 & 21.31 \\
 SegConv   & \textbf{20.86} & \textbf{20.91} & \textbf{20.88}\\
 LogConv   & 21.84 & 21.08 & 22.86 \\
 MixConv & 21.02 & 21.28 & 21.81  \\
 \hline
\end{tabular}
\caption{\footnotesize Testing error rate (\%) of ResNet-32 on CIFAR-100.}
\label{table:cifar100-res}
\end{table}

\par
\vspace{0.5mm}
\setlength{\columnsep}{0.0pt}
\begin{wrapfigure}{r}{0.25\textwidth}
\renewcommand{\captionlabelfont}{\footnotesize}
  \begin{center}
  \advance\leftskip+1mm
  \renewcommand{\captionlabelfont}{\footnotesize}
    \vspace{-0.25in}
    \includegraphics[width=0.23\textwidth]{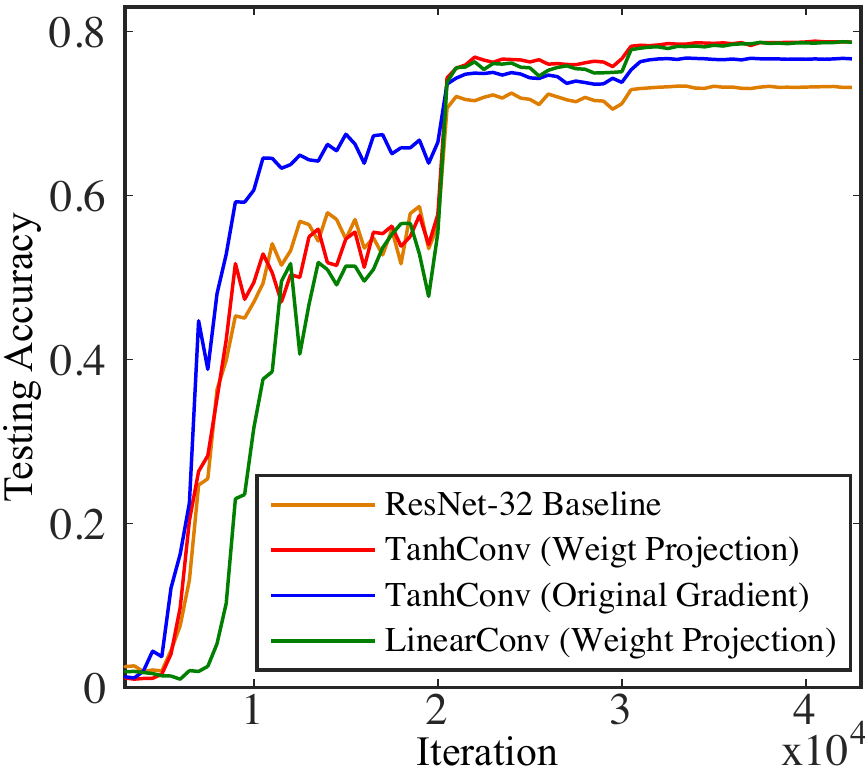}
    \vspace{-0.15in}
    \caption{\footnotesize Convergence.}\label{convergence_mini}
    \vspace{-0.3in}
  \end{center}
\end{wrapfigure}
\noindent\textbf{Convergence.} We also evaluate the convergence of DCNets using the architecture of ResNet-32. The convergence curves in Fig.~\ref{convergence_mini} show that the decoupled operators are able to converge and generalize better than original convolution operators on CIFAR-100 dataset.
\par

\begin{table}[h]
\vspace{-0.6mm}
\footnotesize
\renewcommand{\captionlabelfont}{\footnotesize}
  \setlength{\abovecaptionskip}{1.7pt}
  \setlength{\belowcaptionskip}{-3pt}
\centering
\begin{tabular}{c || c  c}
  \hline
 Method & CIFAR-10 & CIFAR-100 \\
 \hline\hline
 ResNet-110-original \cite{he2016deep}     & 6.61 & 25.16 \\
 ResNet-1001 \cite{he2016identity}  & 4.92 & \textbf{22.71} \\
 ResNet-1001 (64 mini-batch size) \cite{he2016identity}  & \textbf{4.64} & - \\
 \hline
 DCNet-32 (TanhConv + Cosine) & \textbf{4.75} & 21.12\\
 DCNet-32 (LinearConv + Sq. Cos.)       & 5.34 & \textbf{20.23}
 \\\hline
\end{tabular}
\caption{\footnotesize Comparison to the state-of-the-art on CIFAR-10 and CIFAR-100.}
\label{table:SOTA}
\end{table}

\vspace{0.5mm}
\noindent\textbf{Comparison to the state-of-the-art.} Table~\ref{table:SOTA} shows that our DCNet-32 has very competitive accuracy compared to ResNet-1001. In order to achieve best accuracy, we use the weight-normalized softmax loss~\cite{liu2017deephyper}. We also find that using SGD further improves the accuracy of DCNets. Experiments on SGD-trained models are provided in Appendix~\textcolor[rgb]{0.7551,0.2333,0.3333}{F}.

\vspace{-3mm}
\subsubsection{ImageNet-2012}
\vspace{-1mm}
\noindent\textbf{Standard ResNet.} We first evaluate the DCNets with the standard ResNet-18. All presented decoupled operators use the cosine angular activation. \cite{liu2017deephyper} shows that SphereConv can perform comparably to the baseline on ImageNet only when the network is wide enough. Using the weight projection and pretrained model initialization, SphereConv is comparable to the baseline even on narrow networks. Most importantly, TanhConv and LinearConv achieve better accuracy than the baseline ResNet. The learned filters of DCNets on ImageNet are also provided in Appendix~\textcolor[rgb]{0.7551,0.2333,0.3333}{E}.
\par
\vspace{0.5mm}
\noindent\textbf{Modified ResNet.} We also evaluate decoupled operators with a modified ResNet, similar to SphereFace networks~\cite{liu2017sphereface}, to better show the advantages of decoupled operators. DCNets can be trained from scratch and outperform the baseline by 1\%. Moreover, DCNets can converge stably in very challenging scenarios. From Table~\ref{table:imagenet}, we observe that DCNets can converge to a decent accuracy without BN, while the baseline model fails to converge without BN.

\begin{table}[h]
\vspace{-2mm}
\footnotesize
  \setlength{\abovecaptionskip}{2pt}
  \setlength{\belowcaptionskip}{-8pt}
\centering
\renewcommand{\captionlabelfont}{\footnotesize}
\begin{tabular}{c || c c c}
  \hline
  Method & \makecell{Standard \\ ResNet-18 \\ w/ BN} & \makecell{Modified \\ ResNet-18 \\ w/ BN} & \makecell{Modified \\ ResNet-18 \\ w/o BN} \\
 \hline\hline
 Baseline    & 12.63           & 12.10 & N/C     \\
 SphereConv  & 12.68*          & 11.55 & 13.30   \\
 LinearConv  & \textbf{11.99}* & 11.50 & N/C     \\
 TanhConv    & 12.47*          & \textbf{11.10} & \textbf{12.79}   \\
 \hline
\end{tabular}
\caption{\footnotesize Center-crop Top-5 error (\%) of standard ResNet-18 and modified ResNet-18 on ImageNet-2012. * indicates we use the pretrained model of original CNN on ImageNet-2012 as initialization (see Section~\ref{finetune}).}
\label{table:imagenet}
\end{table}

\vspace{-2mm}
\subsection{Robustness against Adversarial attacks}
\vspace{-0.5mm}
We evaluate the robustness of DCNets. DCNets in this subsection use standard gradients and are trained without any optimization trick. Fast gradient sign method (FGSM) \cite{goodfellow2014explaining} and basic iterative method (BIM)~\cite{kurakin2016adversarial} are used to attack the networks. Experimental details and more experiments are given in Appendix~\textcolor[rgb]{0.7551,0.2333,0.3333}{A} and Appendix~\textcolor[rgb]{0.7551,0.2333,0.3333}{C}, respectively.
\vspace{-3.5mm}
\subsubsection{White-box Adversarial Attacks}
\vspace{-2mm}
We run white-box attacks on both naturally trained models and FGSM-trained models on CIFAR-10 (results shown in Table \ref{table:white-box}). ``None'' attacks mean that all the testing samples are normal. For naturally trained models, all DCNet variants show significantly better robustness over the baseline, with naturally trained TanhConv being most resistant. With adversarial training, while DCNets achieve the best robustness, SphereConv is particularly resistant against BIM attack. We speculate that the tight spherical constraint strongly twists the data manifold so that the adversarial gradient updates can only result in small gains.

\begin{table}[h!]
\vspace{-1mm}
\footnotesize
\setlength{\abovecaptionskip}{2pt}
 \setlength{\belowcaptionskip}{-10pt}
\renewcommand{\captionlabelfont}{\footnotesize}
\centering
\begin{tabular}{c||c c c c}
  \hline
  & \multicolumn{4}{c}{Target models} \\
 \makecell{Attack} & Baseline & SphereConv & BallConv & TanhConv \\

 \hline\hline
 None & 85.35 & 88.58 & 91.13 & \textbf{91.45} \\
 FGSM & 18.82 & 43.64 & 50.47 & \textbf{52.60} \\
 BIM  & 8.67  & 8.89  & 7.74 & \textbf{10.18} \\
 \hline\hline
 None & 83.70 & 87.41 & 87.47 & \textbf{87.54} \\
 FGSM & 78.96 & \textbf{85.98} & 82.20 & 81.46 \\
 BIM  & 7.96  & \textbf{35.07} & 17.38 & 19.86 \\
 \hline
\end{tabular}
\caption{\footnotesize White-box attacks on CIFAR-10. Performance is measured in accuracy (\%). The first three rows are results of naturally trained models, and the last three rows are results of adversarially trained models.}
\label{table:white-box}
\end{table}

\begin{table}[h!]
\vspace{-1mm}
\footnotesize
\setlength{\abovecaptionskip}{2pt}
 \setlength{\belowcaptionskip}{-10pt}
\renewcommand{\captionlabelfont}{\footnotesize}
\centering
\begin{tabular}{c||c c c c}
  \hline
  & \multicolumn{4}{c}{Target models} \\
 \makecell{Attack} & Baseline & SphereConv & BallConv & TanhConv \\

 \hline\hline
 None & 85.35 & 88.58 & 91.13 & \textbf{91.45} \\
 FGSM & 50.90 & \textbf{56.71} & 49.50 & 50.61 \\
 BIM  & 36.22 & \textbf{43.10} & 27.48 & 29.06 \\
 \hline\hline
 None & 83.70 & 87.41 & 87.47 & \textbf{87.54} \\
 FGSM & 77.57 & 76.29 & 78.67 & \textbf{80.38} \\
 BIM  & 78.55 & 77.79 & 80.59 & \textbf{82.47} \\
 \hline
\end{tabular}
\caption{\footnotesize Black-box attacks on CIFAR-10. Performance is measured in accuracy (\%). The first three rows are results of naturally trained models, and the last three rows are results of adversarially trained models.}
\label{table:black-box}
\end{table}
\vspace{-3.2mm}
\subsubsection{Black-box Adversarial Attacks}
\vspace{-1.2mm}
We run black-box attacks on naturally-trained and FGSM-trained models on CIFAR-10 (see Table \ref{table:black-box}). With natural training, it is surprising that BallConv and TanhConv do not show an advantage over the baseline, while SphereConv performs the best. The strongly nonlinear landscape of BallConv and TanhConv may be too difficult to be optimized without adversarial training. SphereConv, with a tighter geometric constraint, is able to withstand adversarial attacks without adversarial training. With adversarial training, SphereConv is less resistant against adversarial attacks than the baseline. BallConv and TanhConv, however, show significant advantage over the baseline. Our observation that adversarial training compromises the robustness of SphereConv matches the conclusion made by \cite{tramer2017ensemble}. Since SphereConv enforces a tight constraint of output vectors, the landscape around some data points will be dramatically changed during adversarial training. BallConv and TanhConv are less constrained and thus can fit adversarial examples without detrimental changes in the landscapes.

\vspace{-1mm}
\section{Concluding Remarks}
\vspace{-1mm}
This paper proposes a decoupled framework for learning neural networks. The decoupled formulation enables us to design or learn better decoupled operators than the original convolution. We argue that standard CNNs do not constitute an optimal decoupled design in general.
\par
\vspace{1mm}
\begin{spacing}{0.75}
{\footnotesize \noindent\textbf{Acknowledgements.} The project was supported in part by NSF IIS-1218749, NSF Award BCS-1524565, NIH BIGDATA 1R01GM108341, NSF CAREER IIS-1350983, NSF IIS-1639792 EAGER, NSF CNS-1704701, ONR N00014-15-1-2340, Intel ISTC, NVIDIA, Amazon AWS.}
\end{spacing}

{\small
\bibliographystyle{ieee}
\bibliography{egbib}
}

\newpage

\begin{appendix}

\begin{onecolumn}

\begin{table*}[t]
	\renewcommand{\captionlabelfont}{\footnotesize}
	\newcommand{\tabincell}[2]{\begin{tabular}{@{}#1@{}}#2\end{tabular}}
	\centering
	\setlength{\abovecaptionskip}{3pt}
	\setlength{\belowcaptionskip}{-10pt}
	\footnotesize
	\begin{tabular}{|c|c|c|c|c|c|}
		\hline
		Layer & Plain CNN-9 & CNN-9 for adversarial attacks & ResNet-32 for CIFAR & Standard ResNet-18 & Modified ResNet-18\\
		\hline\hline
		Conv0.x & N/A & N/A & [3$\times$3, 96] & \tabincell{c}{[7$\times$7, 64], S2\\3$\times$3, Max Pooling, S2} & \tabincell{c}{[7$\times$7, 128], S2\\3$\times$3, Max Pooling, S2}\\\hline
		Conv1.x & \tabincell{c}{[3$\times$3, 64]$\times$3\\2$\times$2 Max Pooling, S2}  & \tabincell{c}{[3$\times$3, 32]$\times$3\\2$\times$2 Max Pooling, S2} & \tabincell{c}{$\left[\begin{aligned}&3\times 3, 96\\&3\times3, 96\end{aligned}\right]\times 5$}  & $\left[\begin{aligned}&3\times 3, 64\\&3\times3, 64\end{aligned}\right]\times 2$ & \tabincell{c}{[3$\times$3, 128]$\times$1, S2\\$\left[\begin{aligned}&3\times 3, 128\\&3\times3, 128\end{aligned}\right]\times 1$}\\\hline
		Conv2.x & \tabincell{c}{[3$\times$3, 128]$\times$3\\2$\times$2 Max Pooling, S2}  & \tabincell{c}{[3$\times$3, 64]$\times$3\\2$\times$2 Max Pooling, S2} & \tabincell{c}{$\left[\begin{aligned}&3\times3, 192\\&3\times3, 192\end{aligned}\right]\times 5$} & $\left[\begin{aligned}&3\times 3, 128\\&3\times3, 128\end{aligned}\right]\times 2$ & \tabincell{c}{[3$\times$3, 256]$\times$1, S2\\$\left[\begin{aligned}&3\times 3, 256\\&3\times3, 256\end{aligned}\right]\times 2$} \\\hline
		Conv3.x & \tabincell{c}{[3$\times$3, 256]$\times$3\\2$\times$2 Max Pooling, S2}  & \tabincell{c}{[3$\times$3, 128]$\times$3\\2$\times$2 Max Pooling, S2} & \tabincell{c}{$\left[\begin{aligned}&3\times3, 384\\&3\times3, 384\end{aligned}\right]\times 5$} & $\left[\begin{aligned}&3\times 3, 256\\&3\times3, 256\end{aligned}\right]\times 2$ & \tabincell{c}{[3$\times$3, 512]$\times$1, S2\\$\left[\begin{aligned}&3\times 3, 512\\&3\times3, 512\end{aligned}\right]\times 3$} \\\hline
		Conv4.x & N/A & N/A & N/A & $\left[\begin{aligned}&3\times 3, 512\\&3\times3, 512\end{aligned}\right]\times 2$ & [3$\times$3, 1024]$\times$1, S2 \\\hline
		Final & 512-dim fully connected & 256-dim fully connected & \multicolumn{3}{c|}{Average Pooling}  \\\hline
	\end{tabular}
	\caption{\footnotesize Our CNN and ResNet architectures with different convolutional layers. Conv0.x, Conv1.x, Conv2.x, Conv3.x and Conv4.x denote convolution units that may contain multiple convolutional layers, and residual units are shown in double-column brackets. Conv1.x, Conv2.x and Conv3.x usually operate on different size feature maps. These networks are essentially similar to \cite{he2016deep}, but with different number of filters in each layer. The downsampling is performed by convolutions with a stride of 2. E.g., [3$\times$3, 64]$\times$4 denotes 4 cascaded convolution layers with 64 filters of size 3$\times$3, and S2 denotes stride 2. }\label{netarch2}
\end{table*}

\section{Experimental Details}\label{appendix:expdetail}

\subsection{General Settings}
The network architectures used in the paper are elaborated in Table~\ref{netarch2}. Due to the limitations of our GPU resources, we mostly conduct experiments based on plain CNN-9 and ResNet-32 for CIFAR and ResNet-18 for ImageNet. For CIFAR-10 and CIFAR-100, we use ADAM for all the networks including the baseline. For ImageNet-2012, we use the SGD with momentum $0.9$ for all the networks. If not specified, we use the batch normalization by default for all the experiments on object recognition. For the experiments against adversarial attacks, we use the plain CNN-9. We do not use the batch normalization for the adversarial attack experiments. All the experiments are implemented using TensorFlow library. We use the same data augmentation protocol for CIFAR-10, CIFAR-100 and ImageNet-2012 as \cite{liu2017deephyper}. For initialization of DCNets and baselines, we follow \cite{he2015delving}. For modified ResNet-18 in ImageNet, we use the same initialization as \cite{liu2017sphereface}.

Since we are already using optimization tricks on $\norm{\bm{w}}$, we propose to replace the orthonormal constraint in \cite{liu2017deephyper} with the proposed orthogonal constraint.

\subsection{Details about FGSM and BIM Attacks}

Recent studies show that neural networks are prone to adversarial attacks \cite{goodfellow2014explaining, moosavi2016universal, moosavi2016deepfool, szegedy2013intriguing}. One of the simplest attacks is FGSM \cite{goodfellow2014explaining}, which computes the adversarial image $\tilde{x}$ of some input image $x$ such that $\thickmuskip=2mu \medmuskip=2mu \norm{x - \tilde{x}}_\infty \leq \epsilon$. FGSM performs one single step gradient descent (with step size $\epsilon$) to decrease the probability of the ground truth label. Formally, $\thickmuskip=2mu \medmuskip=2mu \tilde{x} = x + \epsilon \textnormal{sign}(\nabla_x J(\theta,x, y))$ where $J$ is the loss function used to train the network, $\theta$ represents the network parameters, $x$ is the input image and $y$ is the ground truth label associated with $x$. We compare our models and ResNet baseline on the performance on adversarial examples.
\par
In addition, we evaluate the performance of DCNets and ResNet baselines on BIM (Basic Iterative Method) attack ~\cite{kurakin2016adversarial}. BIM runs certain number $N$ of iterations of FGSM, with a smaller step size $\tau$. In each iteration, the resulted perturbed image $\tilde{x}$ is clipped so that $\norm{x - \tilde{x}}_\infty \leq \epsilon$.

We implement the experiments with Cleverhans \cite{papernot2017cleverhans}. In adversarial training using FGSM, $\epsilon=8$ is used to generated the adversarial examples. In all the following adversarial attack experiments, we set $\epsilon = 8$, $\tau = 2$, $N = 20$. We report the accuracy on adversarial examples for both naturally trained models and adversarially trained models using FGSM. The network architecture is shown in Table~\ref{netarch2}.

\subsection{Details about the Black-box Attacks}
\cite{tramer2017ensemble} shows that adversarially trained models behave significantly different on adversarial examples trained on itself and transferred adversarial examples. As suggested by \cite{tramer2017ensemble}, we report the resistance of our models against black-box attacks with ResNet baseline model. Specifically, the adversarial examples are computed from a CNN baseline with the same architecture as the target models. The generated adversarial examples are then used to attack those target models. The architecture and the attack parameters are kept the same as in the white-box experiment.

\section{Training and Implementation Details}\label{appendix:traindetail}

\textbf{Improved Learning of Operator Radius.} To facilitate the learning of the operator radius $\rho$, we multiply the average norm of local patch $\bm{x}$ to $\rho$. Taking TanhConv as an example, we implement it using the following form:
    \begin{equation}\label{tanhim}
    \small
        f_d(\bm{w},\bm{x}) = \alpha\tanh\big(\frac{\norm{\bm{x}}}{\rho\cdot\mathop{\mathbb{E}} \{\norm{\bm{x}}\}}\big)\cdot g(\theta_{(\bm{w},\bm{x})}),
    \end{equation}
where $\rho$ is learnable and it is initialized by a constant $1$. The reason we are multiplying the average norm of $\|\bm{x}\|$ to $\rho$ comes from our empirical observation that lots of $\rho$ in the middle layers stay unchanged and can not be updated effectively. Compared to the original formulation, Eq.~\eqref{tanhim} essentially performs an normalization to $\|\bm{x}\|$ and make its mean become $1$ (\emph{i.e.}, $\mathop{\mathbb{E}}\big(\frac{\norm{\bm{x}}}{\mathop{\mathbb{E}} \{\norm{\bm{x}}\}}\big)=1$). This is also the reason we initialize $\rho$ with 1. The gradient of the magnitude function $h(\cdot)$ w.r.t $\rho$ can be large enough such that $\rho$ is updated effectively. Therefore, for all the decoupled operators that have a learnable non-zero operator radius $\rho$ (\emph{e.g.}, BallConv, TanhConv, SegConv, etc.), we will multiply $\mathop{\mathbb{E}} \{\norm{\bm{x}}\}$ to $\rho$ in order to facilitate its learning. In practice, we use the moving average to compute $\mathop{\mathbb{E}} \{\norm{\bm{x}}\}$, similar to BN~\cite{ioffe2015batch}. Note that, each kernel will preserve its independent patch norm mean $\mathop{\mathbb{E}} \{\norm{\bm{x}}\}$. BallConv is implemented using
    \begin{equation}\label{ballim}
    \small
        f_d(\bm{w},\bm{x}) = \alpha\cdot\frac{\min(\norm{\bm{x}},\rho\cdot\mathop{\mathbb{E}} \{\norm{\bm{x}}\})}{\rho\cdot\mathop{\mathbb{E}} \{\norm{\bm{x}}\}}\cdot g(\theta_{(\bm{w},\bm{x})}),
    \end{equation}
and SegConv is implemented using
\begin{equation}
\small
f_d(\bm{w},\bm{x})=
\left\{
{\begin{array}{*{20}{l}}
{\alpha\norm{\bm{x}}\cdot g(\theta_{(\bm{w},\bm{x})}),\ \ \ \ 0\leq\norm{\bm{x}}\leq\rho\cdot\mathop{\mathbb{E}} \{\norm{\bm{x}}\}}\\
{(\beta\norm{\bm{x}}+\alpha\rho\cdot\mathop{\mathbb{E}} \{\norm{\bm{x}}\}-\beta\rho\cdot\mathop{\mathbb{E}} \{\norm{\bm{x}}\})\cdot g(\theta_{(\bm{w},\bm{x})}),\ \ \ \ \rho\cdot\mathop{\mathbb{E}} \{\norm{\bm{x}}\}<\norm{\bm{x}}}
\end{array}} \right..
\end{equation}

\par
\textbf{Hyperparameter Settings.} For SphereConv, we use $\alpha=1$. For BallConv, we use $\alpha=1$ and a learnable $\rho$. For TanhConv, we use $\alpha=1$ and a learnable $\rho$. For LinearConv, we use $\alpha=1$. For SegConv, we use $\alpha=1$, $\beta=0.5$ and a learnable $\rho$. For LogConv, we use $\alpha=1$. For MixConv, we use $\alpha=1$ and $\beta=1$. For CIFAR experiments, we use 128 batch size for all the networks. For ImageNet experiments, we use 40 batch size for all the networks.

\section{More Experiments on Defense against Adversarial Attacks}
We also evaluate the robustness of DCNets with the DeepFool attacks \cite{moosavi2016deepfool}. Note that, for all the experiments related to the adversarial attacks, our network do not use any optimization trick and is trained by original gradients. The results are given in Fig.~\ref{deepfool}. The x-axis denotes the index of the 10000 adversarial testing samples, and the y-axis denotes the strength ($\ell_2$ norm or $\ell_\infty$ norm) of the perturbations in order to successfully fool the network. We could observe from the results that in order to fool the DCNets, the DeepFool attacks need to largely perturb the samples while it only takes a much smaller perturbation to fool the original CNNs. It implies that DCNets are much more difficult to fool. In other words, to fool the DCNets will take much more efforts than to fool the original CNNs, which shows the superior robustness of DCNets against adversarial examples.
\begin{figure*}[h]
  \centering
  \renewcommand{\captionlabelfont}{\footnotesize}
  \setlength{\abovecaptionskip}{3pt}
  \setlength{\belowcaptionskip}{-10pt}
  \includegraphics[width=6.8in]{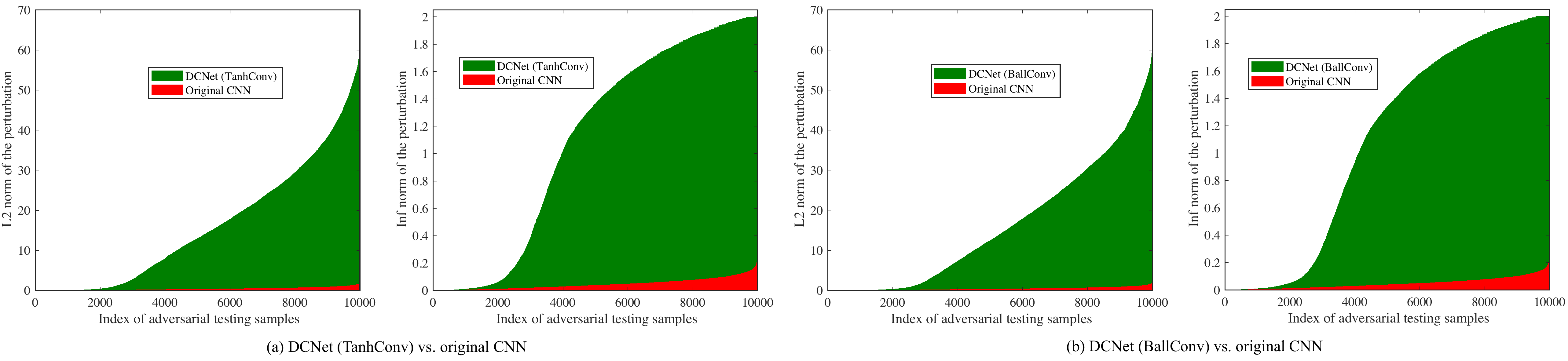}
  \caption{\footnotesize The strength of the adversarial perturbations to fool the network. }\label{deepfool}
\end{figure*}

\begin{figure*}[t]
  \centering
  \renewcommand{\captionlabelfont}{\footnotesize}
  \setlength{\abovecaptionskip}{3pt}
  \setlength{\belowcaptionskip}{-5pt}
  \includegraphics[width=6.8in]{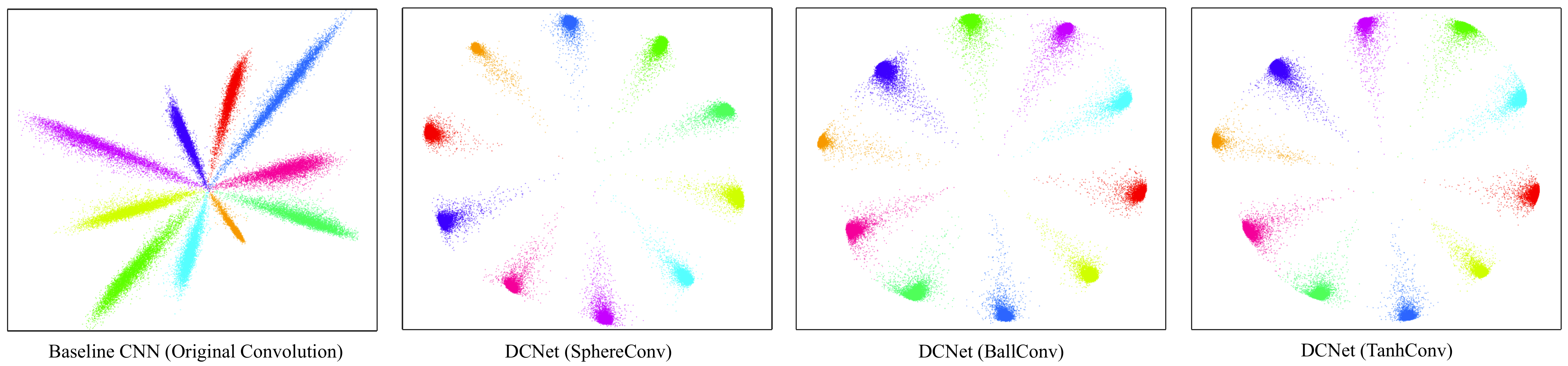}
  \caption{\footnotesize 2D feature visualization on MNIST dataset with natural training. }\label{2d_mnist_feature_natural}
\end{figure*}

\begin{figure*}[t]
  \centering
  \renewcommand{\captionlabelfont}{\footnotesize}
  \setlength{\abovecaptionskip}{3pt}
  \setlength{\belowcaptionskip}{-5pt}
  \includegraphics[width=6.8in]{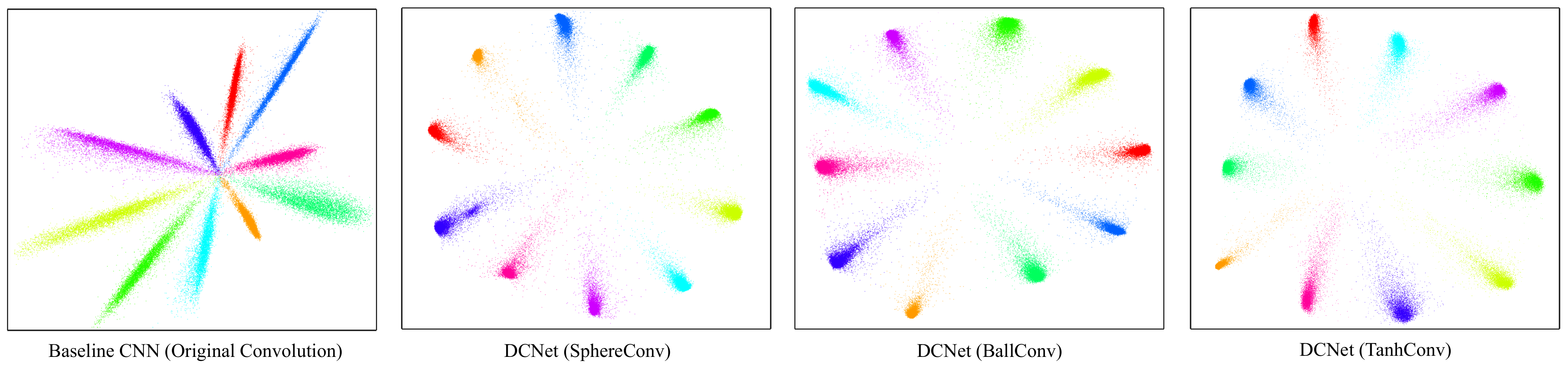}
  \caption{\footnotesize 2D feature visualization on MNIST dataset with adversarial training. }\label{2d_mnist_feature_adversarial}
\end{figure*}

\section{Feature Visualization on MNIST Dataset}
We visualize the 2D feature on MNIST dataset. Specifically, we use a plain CNN with 6 convolutional layers ([3$\times$3,32]$\times$2-[3$\times$3,64]$\times$2-[3$\times$3,128]$\times$2) and 3 fully connected layers (256-2-10). Note that we set the output dimension (\emph{i.e.}, the input dimension of the last fully connected layer) as 2 and visualize these 2D features. We evaluate two types of training: natural training (\emph{i.e.}, trained on the normal MNIST dataset) and adversarial training~\cite{goodfellow2014explaining}. Note that, all the networks in this section are learned by original gradient updates. We do not use weight projection in the networks for the visualization purpose.
\par
\textbf{Natural Training}. We plot the 2D features in Fig.~\ref{2d_mnist_feature_natural}. We could observe that DCNets (especially bounded decoupled operators) exhibit very different distributions with the original CNNs. Empirically, we observe that SphereConv, BallConv and TanhConv produce very compact and well-grouped features.
\par
\textbf{Adversarial Training}. We also show the 2D features of adversarially trained models of baseline CNN, DCNet (SphereConv), DCNet (BallConv) and DCNet (TanhConv) in Fig.\ref{2d_mnist_feature_adversarial}. We could see that DCNets are still able to group the features in a more compact way than the original CNN even with the adversarial training.

\section{Filter Visualization on ImageNet-2012}
We train larger models with 256 filters in the first layer on ImageNet-2012. We visualize all the filters in the first layer for these compared methods in Fig.~\ref{imagenet_filter_all}. One could see that DCNet with SphereConv learns more sparse filters, while DCNets with TanhConv and BallConv can learn richer types of filters. Moreover, because we use orthogonality constraints, the filters are not highly correlated unlike the original CNNs.

\begin{figure*}[h]
  \centering
  \renewcommand{\captionlabelfont}{\footnotesize}
  \setlength{\abovecaptionskip}{3pt}
  \setlength{\belowcaptionskip}{-5pt}
  \includegraphics[width=4.5in]{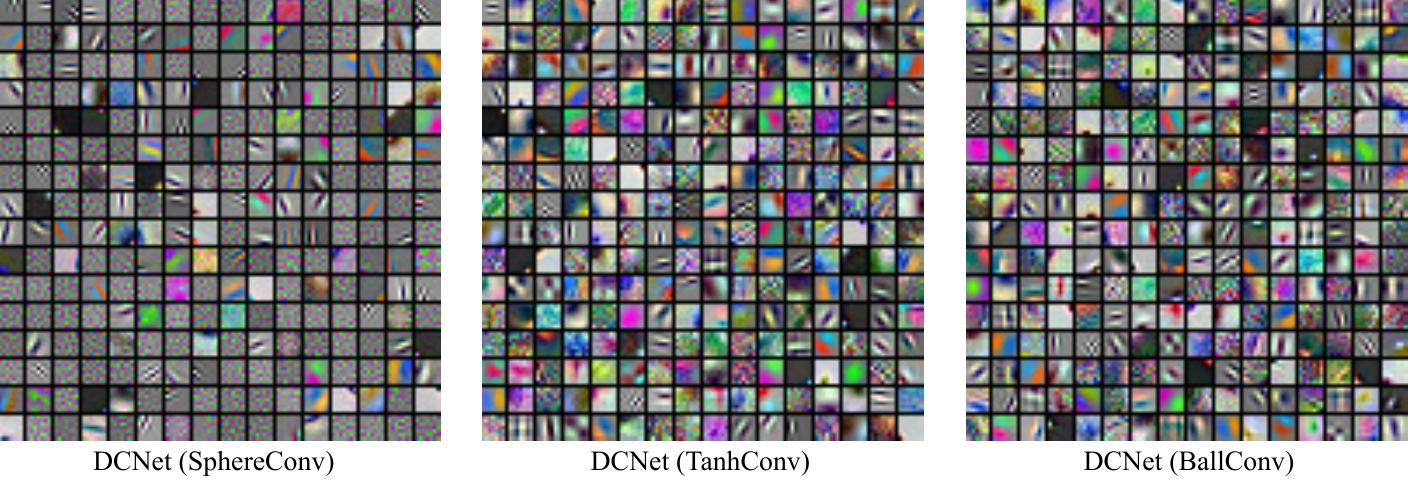}
  \caption{\footnotesize Visualized filters from the first layer of DCNets on ImageNet-2012 dataset. Note that, this is learned by original gradient updates. We do not use weight projection in the networks for the visualization purpose.}\label{imagenet_filter_all}
\end{figure*}

\section{Experiments on CIFAR-100 (stochastic gradient descent)}
Additionally, we optimize the DCNets with stochastic gradient descent (SGD) with momentum and evaluate our models on CIFAR-100. We use the ResNet-32 architecture. The experimental setting is the same as the CIFAR-100 experiment in the main paper, except that we use SGD instead of ADAM. the results are given in Table~\ref{table:cifar100-res-sgd}. Surprisingly, using SGD could largely improve the performance of baseline. While optimizing the baseline model using ADAM gives us 26.69\% error rate, optimizing the baseline using SGD gives us 21.55\% error rate. Even though the baseline will be greatly improved by SGD, we still find that our DCNets optimized by SGD are better than the baseline and also perform slightly better than the DCNets optimized by ADAM.
\begin{table}[h]
\vspace{-1.5mm}
\footnotesize
  \setlength{\abovecaptionskip}{2pt}
  \setlength{\belowcaptionskip}{-8pt}
\centering
\renewcommand{\captionlabelfont}{\footnotesize}
\begin{tabular}{c || c c c}
  \hline
  Method & Linear & Cosine & Sq. Cosine \\
 \hline\hline
 ResNet Baseline      &  -              & 21.55               &  -     \\
 SphereConv           &  21.71          & 21.61          &  24.62 \\
 BallConv             &  20.96          & 21.25          &  24.40 \\
 TanhConv             &  21.07          & 21.12          &  24.29 \\
 LinearConv           &  21.43          & 21.25          &  \textbf{20.54} \\
 SegConv              &  \textbf{20.58} & \textbf{20.61} &  20.61 \\
 LogConv              &  21.15          & 21.42          &  23.10 \\
 MixConv              &  20.82          & 21.20          &  21.19 \\
 \hline
\end{tabular}
\caption{\footnotesize Testing error rate (\%) of SGD-trained ResNet-32 on CIFAR-100.}
\label{table:cifar100-res-sgd}
\end{table}

\section{Difference between Weighted Gradients and Weight Projection} \label{diff_wproj}

\begin{figure*}[h]
  \centering
  \renewcommand{\captionlabelfont}{\footnotesize}
  \setlength{\abovecaptionskip}{3pt}
  \setlength{\belowcaptionskip}{-5pt}
  \includegraphics[width=2in]{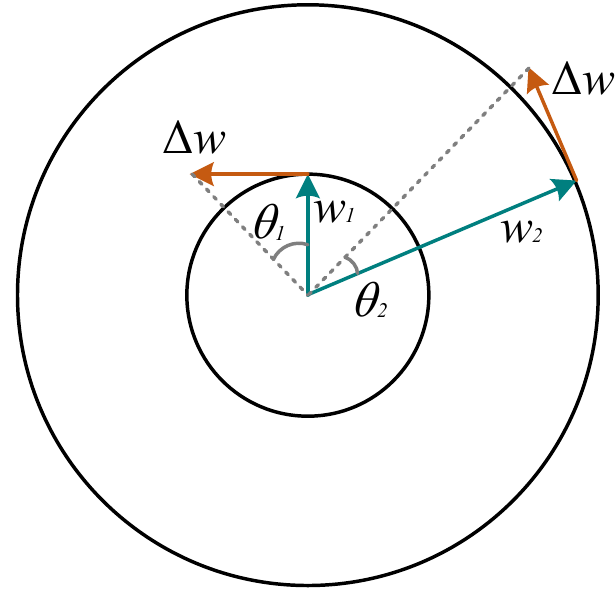}
  \caption{\footnotesize Illustration of weight update, given fixed $||\Delta w ||$. Notice that with $||w_1|| < ||w_2||$, $\theta_1 > \theta_2$.}\label{weighted-radius}
\end{figure*}

It is important to point out that the difference between weight projection and weighted gradients. Although weighted gradients eliminate the effect of normalizing $w$ on the norm of gradients, the increment in angle (i.e. $\theta_{(w, x)}$) is different from that of weighted projection. Consider the simple case of LinearConv. Denote $y = \langle \frac{w}{||w||},x \rangle$. Obviously, the gradient $||\nabla_w y||$ is always perpendicular to $w$. Therefore, \textbf{the original gradient update is optimizing the angle between $w$ and $x$}. The modified gradient update $\Delta w$ is $\nabla_w y \cdot ||w||$ if using weight gradients, and $\nabla_w y$ if using weight projection.

In the case of weighted gradients, even if all updates $\Delta w$ have the same norm, the increment in `angle'  $\Delta \theta = \theta_{(w - \alpha \Delta w, x)} - \theta_{(w, x)}$ can vary, where $\alpha$ is the learning rate. Suppose the norm of the modified gradient $\| \Delta w\| = \|\nabla_w y\| \cdot \|w\|$ is fixed. $\Delta \theta$ can be ignored if $||w||$ is large, while close to $90$ degrees if $\|w\|$ is extremely small. In other words, even if $\Delta w$ is not dependent on $\| w \|$, $\Delta \theta$ is. See Figure~\ref{weighted-radius} for illustration.

In contrast, weighted projection forces the norm of the weights $\|\bm{w}\|$ to be a constant $s$, so when we have fixed gradient $\|\Delta w\|$, the change of angle $\Delta \theta$ will also be a constant (because $\bm{w}$ and $\Delta\bm{w}$ are always perpendicular to each other).

To summarize, weighted gradients make the update of angle $\Delta \theta$ more ``adaptive'', while weight projection makes the update of angle $\Delta \theta$ more ``fixed''. The major reason for such difference is that the norm of the weights $\|\bm{w}\|$ is fixed to a constant $s$ in weight projection, while the norm of the weights $\|\bm{w}\|$ is not a constant in weighted gradients. Different $\|\bm{w}\|$ refers to a hypersphere with different radius, as shown in Fig.~\ref{weighted-radius}.

\end{onecolumn}

\end{appendix}

\end{document}